# SwiftDossier: Tailored Automatic Dossier for Drug Discovery with LLMs and Agents

Gabriele Fossi[a], Youssef Boulaimen[a], Leila Outemzabet[a], Nathalie Jeanray[a], Stéphane Gerart[a], Sébastien Vachenc[a], Joanna Giemza[a], Salvatore Raieli[a,b]

[a]*Oncodesign Precision Medicine, 18 rue Jean Mazen, Dijon, 21000, France*
[b]*Corresponding author: sraieli@oncodesign.com*

## Abstract

The advancement of artificial intelligence algorithms has expanded their application to several fields such as the biomedical domain. Artificial intelligence systems, including Large Language Models (LLMs), can be particularly advantageous in drug discovery, which is a very long and expensive process. However, LLMs by themselves lack in-depth knowledge about specific domains and can generate factually incorrect information. Moreover, they are not able to perform more complex actions that imply the usage of external tools. Our work is focused on these two issues. Firstly, we show how the implementation of an advanced RAG system can help the LLM to generate more accurate answers to drug-discovery-related questions. The results show that the answers generated by the LLM with the RAG system surpass in quality the answers produced by the model without RAG. Secondly, we show how to create an automatic target dossier using LLMs and incorporating them with external tools that they can use to execute more intricate tasks to gather data such as accessing databases and executing code. The result is a production-ready target dossier containing the acquired information summarized into a PDF and a PowerPoint presentation.

*Keywords:* LLM Agents, Automatic Target Dossier, LLMs in Biology, Drug Discovery

## 1. Introduction

Drug discovery is an expensive, lengthy, and high-risk process. The process can cost up to $ 1-2 billion and takes an average of 10-15 years [19]. The failure rate of a drug candidate in clinical trials reaches 90 % (with a higher rate if preclinical stages are also considered). Among the main causes that can be identified are: lack of clinical efficacy, toxicity, and poor pharmacokinetic and pharmacodynamic properties [9, 8, 5]. In recent years, the development of new artificial intelligence algorithms has been extended to various fields, including biomedical science. The large availability of available data means that the pharmaceutical sector can be a perfect playground for AI [10, 17]. The development of these algorithms promises to revolutionize drug discovery and ameliorate the process from a target to a drug on the market. Drug discovery is a complex, multi-step process; AI looks promising in most steps [4, 1, 18]. At the same time, a cautionary approach is required because the capabilities of an AI model depend heavily on the data with which it is fed. For these promises to be fulfilled, AI approaches must be rigorous and based on quality data [30]. Large Language Models (LLMs) have become the main direction of artificial intelligence in recent years and have shown remarkable generalist skills. LLMs have also shown promising capabilities for different scientific tasks and also in the biomedical field [54, 25]. There has been considerable effort to adapt these models to the needs of the medical and at the same time pharmaceutical fields [11, 50]. One of the main problems with the use of LLMs in the pharmaceutical field (and more broadly in the biomedical field) is the generation of hallucinations by models [32]. LLMs can generate outputs that are inaccurate or that contain factually incorrect information. At the same time, they can generate extrinsic hallucinations (claims that cannot be verified) such as quotes that cannot be verified. Another problem that affects their use is that these models are generalist by nature and are not often adapted to the domain of interest [34]. In addition, the architecture of LLMs (transformer derivative) does not easily and not expensively allow ingesting new knowledge (as well as making it difficult to correct previously acquired erroneous knowledge ) [29, 27]. The first step in the target discovery process is generally target identification and selection. Once a target has been identified, the target dossier is one of the most widely used tools for evaluating the goodness-of-fit of a target [6, 20]. The target dossier comprehensively presents all the information that experts need in the decision-making process. A target dossier is an essential tool that must present information about the target, its impact on the disease, pros and cons, therapeutic opportunities, and potential competitors. The dossier preparation process requires extensive time in researching and analyzing different sources. Automating it will save considerable time and help in standardization. For appropriate decision-making, the target dossier must contain correct, up-to-date, and verifiable information. For this reason, it is not possible today to use a classical LLM alone for this task.

In this technical report, we show how it is possible to use an LLM to create an automatic target dossier. Our system that employs both Agents and retrieval augmented generation (RAG) allows us to overcome the limitations highlighted above. The system, given a target (gene name) and a pathology of interest, automatically generates a target dossier (a summary PDF document) and a PowerPoint presentation. The system retrieves up-to-date information from various databases and allows verifiability of the information by annotating the sources used. Our main contributions are:

- we show how a modified pipeline of RAG dramatically increases the quality of response in the biomedical domain.

- we show how using an LLM is possible to create an automatic target dossier. The system accesses article databases, external databases, and tools, executing code and plot autonomously. In addition, it produces a high-quality summary PDF and a ready-to-use PowerPoint presentation.

A complete example of the generated target dossier (both PDF and PowerPoint) is present at https://github.com/Oncodesign-Precision-Medicine/Automatic-Target-Dossier

## 2. Related Work

### 2.1. LLM hallucinations

Hallucinations are considered among the most important challenges for LLM deployment in biomedical or drug discovery [54]. Hallucinations can occur when models produce texts that include details, facts, or claims that are either fictional or misleading [32]. In addition, LLMs are capable today of producing content that sounds plausible and based on scientific patterns they learned during training. In some cases, it has been seen that an LLM is capable of generating pseudo-citations that can further confuse the reader. Hallucinations are a real problem, especially for those disciplines where a pattern must necessarily be factually correct (e.g. healthcare) [54]. It is not easy to be able to correct this behavior because there are different types of hallucinations each with probably a different cause [41]. For this reason, several approaches have been developed to mitigate hallucinations in the outputs produced by LLMs [51]. Some techniques are based on the idea of providing the model with the necessary context to be able to answer a question. Retrieval-augmented generation (RAG) allows the LLM to access authoritative, external information, especially when the answer is not present in the parametric memory of the model [46, 43]. In contrast to other methods such as fine-tuning the computational cost is much lower and it is possible to integrate new information easily [40]. However, naive RAG often has limitations (pre-retrieval, retrieval, and post-retrieval issues) [43] so numerous add-ons have evolved [43]. At the same time, finding an effective strategy to create an advanced pipeline of RAGs is an expensive and time-consuming process. One of the sore points of RAG is especially to make sure that all relevant documents are found. In the biomedical field, the exact answer can be directly found in specialized databases. For this in recent times, Agents have been used where complex reasoning is required, thus allowing LLMs to control different tools and query different databases via API [53]. In this work, we use both RAG and Agents to interact with databases.

### 2.2. Domain adaptation and continuous learning with LLMs

LLMs acquire most of their knowledge during pretraining. Once pretraining is over, it is difficult to ingest new knowledge within the model. This can be done by fine-tuning. However, this is an expensive and inefficient process that would have to be repeated every time the model update is to be conducted [37]. In addition, LLM struggles to integrate this new knowledge. Recent studies show that fine-tuning increases the model's tendency to hallucinate [44]. Second, the LLM tends to forget the old knowledge, and there arises at the same time the discrepancy between the remaining old knowledge and the new knowledge with the risk of bias and unintended association [27]. RAG and Agents present a possible solution to these problems. It is not necessary to conduct parameter editing of the LLM but the new knowledge needed is found in real-time and provided in context to the model. Moreover, in several databases the information is rigorously cataloged and annotated, allowing filtering of irrelevant information.

### 2.3. LLM for target dossier

To the best of our knowledge, there is currently no system that creates an automatic target dossier. LLMs have been proposed for various uses in drug discovery pipelines. For example for molecule optimization, drug-drug interaction, molecular property predictions [38, 28, 52]. On the other hand, in several biomedical tasks, generalist LLMs perform poorly [47]. There are, however, studies that show that it is effective to extend the capabilities of models with the use of tools. ToolLLM and HuggingGPT are examples where an LLM is connected with different APIs or can call other models to perform different tasks [33, 31]. GeneGPT connects an LLM with external APIs from the National Center for Biotechnology Information (NCBI) [45]. The system shows superior capabilities in GeneTuring benchmark vs. other generalist or trained models on biomedical tasks. MedAgents is another framework that shows the superior capabilities of an agent-based system for medical tasks [35]. The capabilities of these systems show the goodness of an approach no longer based only on one LLM but with the integration of other tools.

## 3. Materials and methods

### 3.1. Data

A target dossier includes a large variety of information, spanning from insights about a specific target gene to details about the disease, and includes sections in which reasoning is crucial. To this end, several data coming from different sources are needed, in order to have a comprehensive view of the target and allow the LLM to perform high-level reasoning. The data sources and tools employed in this work are listed in Table 1. The specific sources used are specified in both the target dossier PDF and PowerPoint as well. We used reference citations to allow us to track back which sources are used for each section and slide. For example, when the LLM is using the RAG, the identifiers (Pubmed ID) of the article in the context are noted in the document. In the PowerPoint in the notes are stored the details of each data used. As an example, when the structure of a protein is used, we note the accession link allowing us to retrieve which information the LLM used.

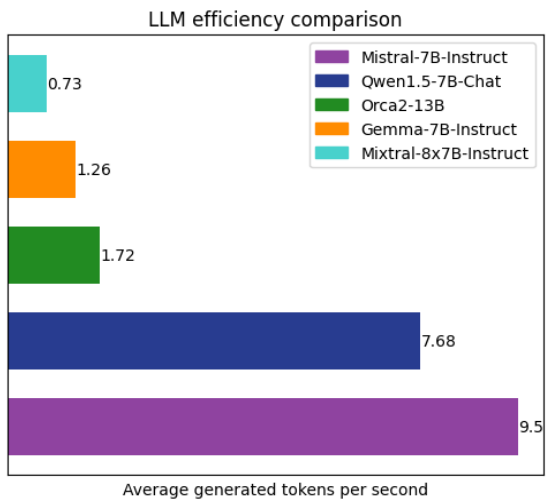

Figure 1: Comparison of LLM efficiency. The height of the bars indicates the average number of generated tokens per second. The value is indicated on the right of the bar.

### 3.2. Models

We decided to use in this work open-source LLMs. Many open-source models differ in architecture and number of parameters [39, 48]. Thus, different LLMs were compared (Mistral-7B-instruct, Qwen-1.5-7B-Chat, Orca 13B, Gemma-7B-instruct, Mistral-8x7B-instruct), focusing on the quality of their answers to a specific set of questions and their efficiency in token generation. Figure 1 shows the results of this analysis. All the assessed models produced good answers, with Mixtral-8x7B-Instruct excelling in this task. However, being a big model, it was slower compared to the other ones. An optimal trade-off between effectiveness and efficiency was Mistral-7B-Instruct [21] which was picked and used in our work.

LLMs contain a substantial amount of information but highly rely on the data that has been used to train them [49]. To expand their internal knowledge by integrating retrieval-based memories, it is possible to leverage Retrieval-Augmented Generation (RAG) [13]. RAG is particularly useful in domain-specific applications, especially in the biomedical domain as described above.

Our RAG framework incorporates reranking and Hybrid Search methodologies, with the embedding model fine-tuned to enhance search effectiveness. Specifically, we utilized *bge-base-en* [36] for this purpose. The model was fine-tuned using a synthetic dataset about pancreatic cancer generated by Qwen 7B [22]. Regarding the reranking model, *mxbai-rerank-large* was selected. Both these models are relatively small, making them suitable to execute RAG dynamically.

The information that is used for RAG and is provided as context for Mistral 7B is gathered from two different sources: PubMed abstracts and PMC articles. The abstracts are short, therefore they do not need any preprocessing and can be directly embedded. On the other hand, articles are too long. Thus they are split into chunks using LangChain *SemanticChunker* before the embedding process. The abstracts and articles are embedded using our fine-tuned *bge-base-en* model and stored in a temporary Chroma DB collection. A similarity search between the query and the embeddings is performed, resulting in a list of documents that are reranked and presented to the LLM as the context to generate a text or answer a question. After the termination of the process, the documents and embeddings in the temporary collection are deleted. Besides the content of the abstracts and articles, their PMIDs are extracted too, to reference the specific documents that have been used by the LLM to generate a response. We evaluated our RAG strategy using GPT4 as evaluator [24]. In this work, we compared the answers to 22 drug discovery questions that were written by experts in the domain. We compared three different possibilities: LLM alone (Mistral 7B), naive RAG, and our RAG pipeline.

### 3.3. System Description

The process begins with the user inputting a query that is related to the generation of a target dossier. Both the target and pathology of interest need to be specified. In this work, we focused on pancreatic cancer, which is a deadly disease that needs therapeutic options [26]. The query is directed to the Agent which has different tools at its disposal. These tools allow the Agent to connect to multiple databases and extract the relevant information. They enable the Agent to:

- Execute Python code to connect to the APIs, process data, and generate plots,

- Retrieve images from the databases,

- Generate texts and summaries using the LLM integrated with the RAG system described previously.

Once all the information has been retrieved, it is aggregated and compiled into the target dossier, producing both a PDF and a PowerPoint presentation. The PDF is generated before the PowerPoint since the latter consists of an encapsulated version of the PDF. The system is summarized in Figure 2.

### 3.4. Technical Requirements

This work revolves around three main platforms/applications: HuggingFace [12], LangChain [16] and Chroma DB [55]. HuggingFace is the platform that we used for model loading and usage since it provides a wide variety of pre-trained models that can be easily accessed. LangChain is the framework that we employed to develop the Agent starting from the LLM and for the creation of the tools. An advantage of using LangChain is that it allows an easy integration with HuggingFace. Finally, as a database to implement our RAG system, we used Chroma DB. It allows the storage of the document embeddings and their retrieval during the RAG process. These tools were used through version 3.11.5 of Python. The system was run on 8 Tesla V100 GPUs, each with 32GB of memory.

| Source | Description | Web Address |
| --- | --- | --- |
| UniProt | A database offers high-quality and freely accessible resource for protein sequence and functional information. | uniprot.org |
| Human Protein Atlas | A database focused on genome-wide analysis of human proteins | proteinatlas.org |
| DrugBank | An online database containing information on drugs and drug targets. | go.drugbank.com |
| Open Targets | A database that uses human genetics and genomics data for drug target identification and prioritization. | opentargets.org |
| RCSB PDB | An online database containing information on drugs and drug targets. | rcsb.org |
| cBioPortal | A resource for the exploration of cancer genomics datasets containing information such as mutation frequencies. | cbioportal.org |
| TCGA Survival | A website that provides analysis of mutations, copy number alterations, etc. Associated with cancer outcome in TCGA. | tcga-survival.com |
| OGEE | An online database containing information about gene essentiality. | v3.ogee.info |
| STRING | A database about protein-protein interactions including both direct and indirect associations. | string-db.org |
| SIGNOR | A repository of annotated causal relationships among human proteins, chemicals of biological relevance, stimuli and phenotypes | signor.uniroma2.it |
| ESMO | The European Society for Medical Oncology provides oncology information, including guidelines on cancer. | esmo.org |
| PubChem | A database that contains information about molecules such as chemical structures, chemical and physical properties and biological activities. | pubchem.ncbi.nlm.nih.gov |
| Gene | An NCBI database that integrates gene information from different species. | ncbi.nlm.nih.gov/gene |
| PubMed | An NCBI database comprises millions of citations and abstracts about biomedical literature. | pubmed.ncbi.nlm.nih.gov |
| PMC | An NCBI archive of biomedical and life sciences journal literature. | ncbi.nlm.nih.gov/pmc |
| BLAST | A tool that finds the similarity between biological sequences | blast.ncbi.nlm.nih.gov/Blast.cgi |
| DeepTMHMM | A Deep Learning Model for classification and prediction of Transmembrane Topology | dtu.biolib.com/DeepTMHMM |
| GSEApy | A Python package to perform Gene Set Enrichment Analysis | github.com/zqfang/GSEApy |

Table 1: List of the sources containing a short description and the web address to access them.

## 4. Results

### 4.1. RAG adaptation to drug discovery domain

To evaluate the effectiveness of RAG, the answers of three different models were compared: the LLM without RAG, the LLM with the base RAG, and the LLM with our RAG system illustrated in Section 3.2. The answers were evaluated by GPT-4 which was specifically prompted to assign a score to each answer according to 4 metrics (faithfulness, relevance, quality, completeness, and correctness). The scores are summarized in the Box Plots of Figure 3. The model with the complete RAG system outperforms the other two models. 75% of its scores are equal to or greater than 4 for every metric. Moreover, its median is 5 and is greater than the median of the other two models which is equal to 4.

One question and the models answers are illustrated in Table 2.

The answer of the model without RAG is entirely incorrect since it fails to understand that pancreatic ductal adenocarcinoma (PDAC) is the subject of the question. Instead, it discusses the Pacific Data and Automation Corporation which has the same acronym PDAC.

### 4.2. Automatic Target Dossier

In this study, we focused on pancreatic cancer, one of the cancers with the most dismal prognosis. To date, pancreatic cancer presents few therapeutic opportunities and is in dire need of new drugs for treatment [14]. To show the system's capabilities, we chose KRAS, one of the genes with the greatest impact on the disease and of interest as a potential target [3, 2, 15, 7]. However, the system is generalizable and can therefore be used with any gene and any pathology. Choosing another gene and another pathology, the system will search for the necessary information. The output of the system is the target dossier, generated as two different files: a PDF and a PowerPoint presentation. The table of contents of the PDF is illustrated in Figure 4. It is divided into four main sections. The first one (containing most of the information) comprises details about the target of interest such as its characteristics, expression, and mutations. The second section is related to the cancer for which we are evaluating the impact of the target and is mainly generated by the LLM using the RAG mechanism. The following section focuses on the existing drugs targeting the gene, with particular attention to the drugs used to treat the cancer of interest. This section is significant from the perspective of competition. Finally, the last section encompasses the conclusions and is the most challenging one for the LLM. Indeed, the SWOT analysis requires high-level reasoning and understanding as well as a highly specific prompt. After the generation of the PDF, the information is summarized and compiled into a PowerPoint presentation.

| | **Question: "Can you describe the PDAC escape mechanisms?"** |
|---|---|
| **Without RAG** | The PDAC (Pacific Data and Automation Corporation) escape mechanisms are a set of security features designed to prevent unauthorized access to the data stored on a PDAC device. These mechanisms include:<br>1. Encryption: All data stored on a PDAC device is encrypted using a proprietary algorithm. This ensures that even if an attacker gains access to the device, they will not be able to read the data without the encryption key.<br>2. Authentication: Before accessing any data on a PDAC device, the user must first authenticate themselves using a valid username and password. This ensures that only authorized users can access the data.<br>3. Access Control: Access to data on a PDAC device is controlled using a set of permissions. Each user is assigned a set of permissions that determine what data they can access and what actions they can perform on that data.<br>4. Audit Trail: All actions performed on a PDAC device are logged in an audit trail. This allows administrators to track who accessed the data and when, and to detect any unauthorized access attempts.<br>5. Physical Security: PDAC devices are designed to be physically secure, with features such as tamper-evident seals and secure locking mechanisms to prevent unauthorized access to the device.<br>Overall, the PDAC escape mechanisms are designed to provide a high level of security for the data stored on a PDAC device, and to prevent unauthorized access to that data. |
| **Base RAG** | The provided context discusses two possible escape mechanisms for Pancreatic ductal adenocarcinoma (PDAC):<br>1. Gasdermin E (GSDME) and mucin expression: PDAC cells use GSDME to form a barrier that prevents chymotrypsin-mediated destruction. GSDME facilitates the expression of mucin 1 and mucin 13, which form this barrier. Inhibition or knockout of mucin 1 or mucin 13 abrogates orthotopic PDAC growth in NOD-SCID mice.<br>2. Autophagy and Major Histocompatibility Complex class I (MHC-I) sequestration: Autophagy was recently described as a tumor-intrinsic immune escape process during tumor development by sequestering MHC-I inside PDAC cells. This discovery implies that autophagy may be a potential target for limiting immune escape in patients and improving immunotherapy efficiency.<br>It is important to note that the context does not provide enough information to fully understand the PDAC escape mechanisms or their implications for treatment. Further research and studies are needed to fully understand and address these mechanisms. |
| **New RAG** | Based on the provided documents, it appears that pancreatic ductal adenocarcinoma (PDAC) has several escape mechanisms that allow it to evade the immune system and survive in the pancreatic microenvironment. One such mechanism involves the use of gasdermin E (GSDME) to form a barrier that prevents chymotrypsin-mediated destruction. GSDME facilitates the expression of mucin 1 and mucin 13, which form this barrier. Inhibition or knockout of mucin 1 or mucin 13 abrogates orthotopic PDAC growth in NOD-SCID mice. Another mechanism involves the sequestration of Major Histocompatibility Complex class I (MHC-I) inside PDAC cells through autophagy. This process supports tumor progression and may be a potential therapeutic target. Additionally, tumor-derived HLA-G may indicate the mechanism of immune escape and impaired PDAC clinical outcome, especially in early-stage patients. Overall, understanding these escape mechanisms is crucial for the development of more effective immunotherapies for PDAC. |

Table 2: Table showing the answers of the three models to the question "Can you describe the PDAC escape mechanisms?"

Both PDF and presentation contain references, specifying which database/tool was used to obtain the information. When available, specific web addresses or IDs are provided, in order to facilitate the reader to access the data and the documents that have been used by the LLM for generation. In the PDF, these specific links can be found at the end in the 'Sources' section. Conversely, in the PowerPoint presentation, they are added in the speaker notes.

Three example slides are illustrated below. Figure 5 shows a table with some characteristics of the target: the similarity between the human protein sequence and the sequences of some animals as well as the protein function. The Agent retrieves the sequences from the UniProt API and they are written in FASTA files, a format which is suitable to represent protein sequences. These files are sent to BLAST which returns the alignment match which can be used to compute the similarity. The protein function is extracted from UniProt and is summarized by the LLM in case it exceeds a certain length. This slide shows the system's ability to retrieve information using external tools, summarize texts using the LLM, and display the data in a table to facilitate readability.

Figure 6 provides insights about the subcellular location of the target in the Human Protein Atlas database. The three pictures illustrating the location are extracted one by one and the names of the represented cell lines are added in the lower-right corner of each image. The antibody used for the analysis is reported in the slide and a legend explaining the colours in the image is added. The image allows the reader to have a clear view of the subcellular location of the target (production-ready). This slide shows the ability of the system to recover images from databases and websites.

Figure 7 illustrates the Agent's ability to execute Python code. Firstly, the list of the one hundred genes with the highest interaction scores with the target is gathered from the STRING database. Afterward, pathway enrichment analysis is performed using the Python library GSEApy (based on the previous obtained list from STRING). Both gene lists and gene sets need to be specified. In our case, as gene sets, we considered Gene Ontology Biological Process 2023, Molecular Function 2023, and Cellular Component 2023 as well as KEGG 2021 Human. Lastly, the results of the pathway enrichment analysis are plotted and the graph is added to the target dossier.

## 5. Conclusion

Our work demonstrates two key findings. Firstly, we showed how the employment of RAG can help increase the quality of the answers in drug discovery applications. Indeed, the LLM without RAG lacks knowledge about the biomedical domain and can generate off-topic answers that are unrelated to the subject of the question. Moreover, the exploitation of a more advanced RAG system determines an additional enhancement in the quality of the responses, due to an improved retrieval process. Regarding the generation of the automatic target dossier, we showed how it is possible to integrate the LLM with some tools that allow it to perform more complex actions such as accessing databases, executing Python code, and compiling all the retrieved information into a PDF and a PowerPoint presentation. We show a system capable of creating production-ready documents and presentations and can have a valuable impact on the work in drug discovery.

Toxicity and lack of clinical efficacy are among the main causes of molecule failure. A detailed target dossier eviscerates the role of a gene in pathology, and potential adversarial effects helping in predicting these critical causes of failure. Indeed, in the target dossier, there are sections devoted to both potential toxicity and the role the gene plays in disease physiology and pathogenicity, aiding in decision-making. So, our second finding is that an AI model can generate a consistent document and presentation when instructed by experts on where to find the information. In response to a user's question, the model automatically generates both the presentation and the PDF. LLMs are capable of generating text, but this is usually plagued by hallucinations, incomplete, and incorrect information. Here we show that an LLM can correct these deficiencies when instructed where to find the information (article repository, biological databases, code to run). The automatic generation of the target dossier can be seen as the first step of a more complex system. An automatic target dossier is an important step in standardizing the process. The automatic target dossier can be a valuable tool in helping different stakeholders in a pharmaceutical company, saving time and helping them to keep up to date quickly. Although the input of domain experts (biologists, physicians, chemists) in the creation of a target dossier remains key, this system allows a presentation and report to be generated automatically. This can be seen as a first step in creating different AI assistants that help the experts' daily work. Different works show how AI is not meant to replace but to support them in daily tasks the experts, increasing their productivity and reducing their workload [42, 23]. Another improvement could be the utilization of a multimodal AI model. We used here an LLM for the reasoning steps, but our system retrieved and plotted images. A multimodal model can conduct reasoning by taking into account additional modalities.

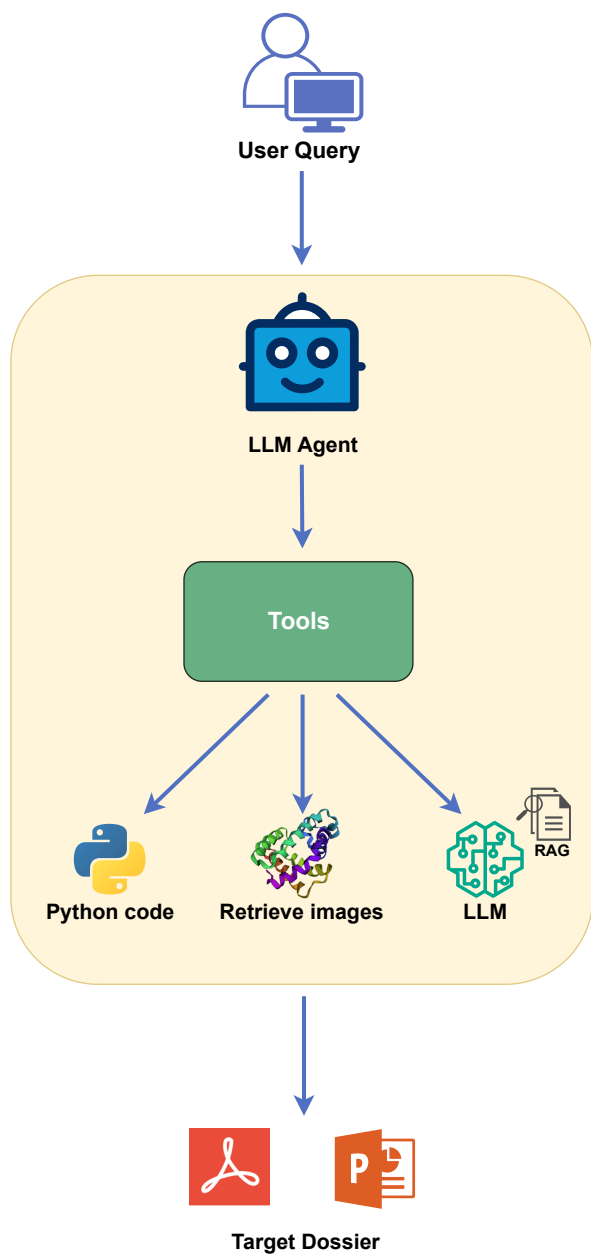

Figure 2: Diagram of the system. The user query is sent to the Agent that uses the available tools to extract information from the databases and generate texts. The information is collected into the target dossier PDF and PowerPoint which are generated and saved.

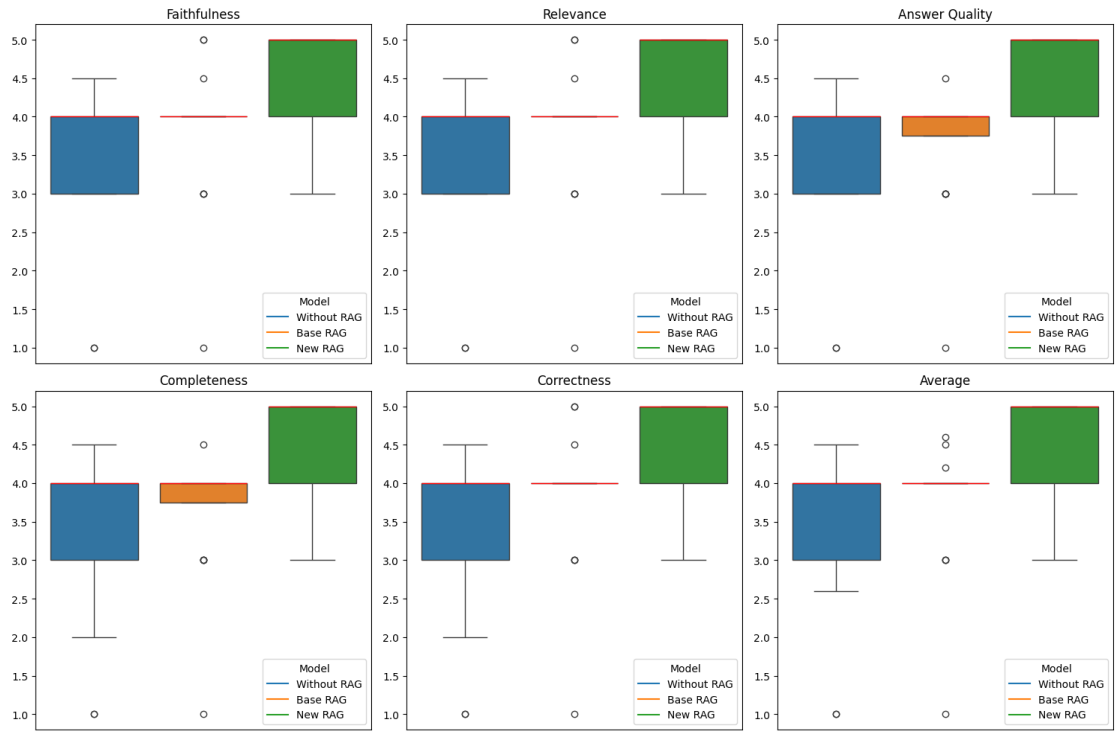

Figure 3: Box Plots summarizing LLM answers evaluation performed by GPT-4. The median is represented by the horizontal red line. The plots show the superior performance of the model with complete RAG.

Figure 4: Table of contents of the PDF target dossier. It is divided into four main sections: Target information, Disease information, Competitive landscape, and a final section for the conclusions.

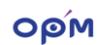

Figure 5: Slide showing some characteristics of the target formatted as a table. The information is collected using UniProt and BLAST.

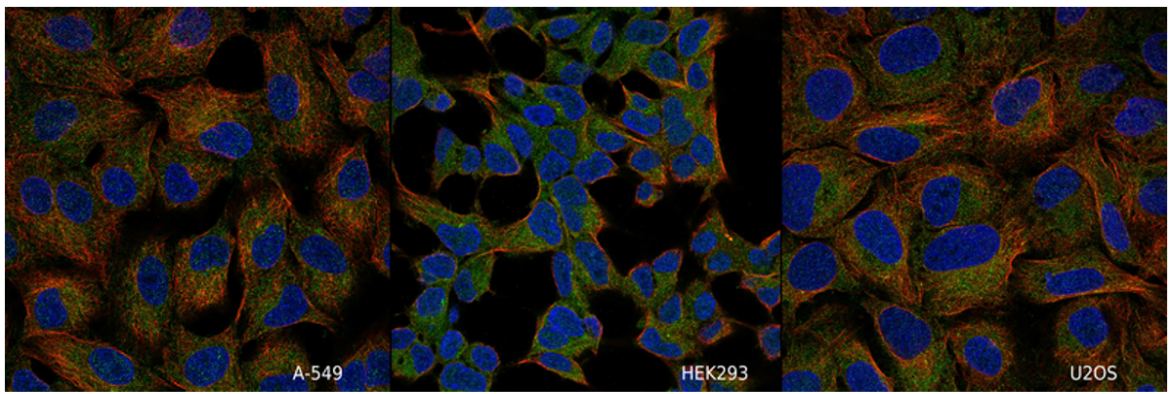

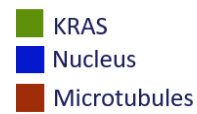

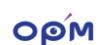

Figure 6: Slide showing information about the subcellular location of the target available in the Human Protein Atlas.

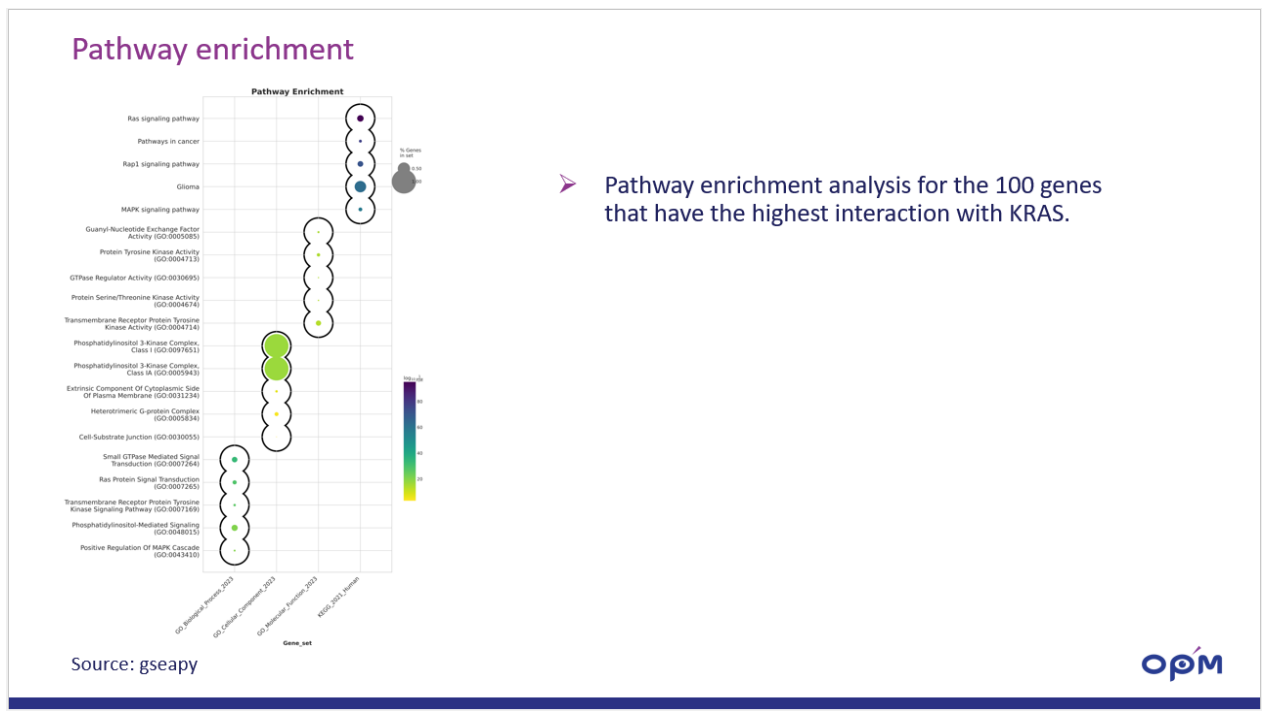

Figure 7: Slide showing the results of the pathway enrichment analysis. The list of genes which interact with the target is retrieved from the STRING database while the enrichment analysis is performed using the Python package GSEApy.

**Acknowledgements**

We would like to thank the members of the IT team, Jonathan Schmiedt and Thomas Wursten for their technical assistance during this project. We would also like to thank Maria Eugenia Riveiro, Kenji Shoji, and Oleksandr Levenets, Jan Hoflack for bringing their invaluable expertise and thorough insights.

**References**

[1] Steven M Paul et al. "How to improve R&D productivity: the pharmaceutical industry's grand challenge". In: *Nature reviews Drug discovery* 9.3 (2010), pp. 203–214.

[2] S Eser et al. "Oncogenic KRAS signalling in pancreatic cancer". In: *British journal of cancer* 111.5 (2014), pp. 817–822.

[3] Barbara Bournet et al. "Targeting KRAS for diagnosis, prognosis, and treatment of pancreatic cancer: Hopes and realities". In: *European journal of cancer* 54 (2016), pp. 75–83.

[4] Joseph A DiMasi, Henry G Grabowski, and Ronald W Hansen. "Innovation in the pharmaceutical industry: new estimates of R&D costs". In: *Journal of health economics* 47 (2016), pp. 20–33.

[5] Richard K Harrison. "Phase II and phase III failures: 2013–2015". In: *Nat Rev Drug Discov* 15.12 (2016), pp. 817–818.

[6] Theodora Katsila et al. "Computational approaches in target identification and drug discovery". In: *Computational and structural biotechnology journal* 14 (2016), pp. 177–184.

[7] Karen M Mann et al. "KRAS-related proteins in pancreatic cancer". In: *Pharmacology & therapeutics* 168 (2016), pp. 29–42.

[8] Tohru Takebe, Ryoka Imai, and Shunsuke Ono. "The current status of drug discovery and development as originated in United States academia: the influence of industrial and academic collaboration on drug discovery and development". In: *Clinical and translational science* 11.6 (2018), pp. 597–606.

[9] Helen Dowden and Jamie Munro. "Trends in clinical success rates and therapeutic focus". In: *Nat Rev Drug Discov* 18.7 (2019), pp. 495–496.

[10] Kit-Kay Mak and Mallikarjuna Rao Pichika. "Artificial intelligence in drug development: present status and future prospects". In: *Drug discovery today* 24.3 (2019), pp. 773–780.

[11] W Patrick Walters and Regina Barzilay. "Applications of deep learning in molecule generation and molecular property prediction". In: *Accounts of chemical research* 54.2 (2020), pp. 263–270.

[12] Thomas Wolf et al. *HuggingFace's Transformers: State-of-the-art Natural Language Processing*. 2020. arXiv: 1910.03771. URL: https://arxiv.org/abs/1910.03771.

[13] Patrick Lewis et al. *Retrieval-Augmented Generation for Knowledge-Intensive NLP Tasks*. 2021. arXiv: 005.11401.

[14] Jie Cai et al. "Advances in the epidemiology of pancreatic cancer: Trends, risk factors, screening, and prognosis". In: *Cancer letters* 520 (2021), pp. 1–11.

[15] Ji Luo. "KRAS mutation in pancreatic cancer". In: *Seminars in oncology*. Vol. 48. 1. Elsevier. 2021, pp. 10–18.

[16] Harrison Chase. *LangChain*. Oct. 2022. URL: https://github.com/langchain-ai/langchain.

[17] Jianyuan Deng et al. "Artificial intelligence in drug discovery: applications and techniques". In: *Briefings in Bioinformatics* 23.1 (2022), bbab430.

[18] Madura KP Jayatunga et al. "AI in small-molecule drug discovery: a coming wave". In: *Nat. Rev. Drug Discov* 21.3 (2022), pp. 175–176.

[19] Duxin Sun et al. "Why 90% of clinical drug development fails and how to improve it?" In: *Acta Pharmaceutica Sinica B* 12.7 (2022), pp. 3049–3062.

[20] Yujie You et al. "Artificial intelligence in cancer target identification and drug discovery". In: *Signal Transduction and Targeted Therapy* 7.1 (2022), p. 156.

[21] Albert Q. Jiang et al. *Mistral 7B*. 2023. arXiv: 2310.06825.

[22] Jinze Bai et al. *Qwen Technical Report*. 2023. arXiv: 2309.16609.

[23] Erik Brynjolfsson, Danielle Li, and Lindsey Raymond. *Generative AI at Work*. 2023. arXiv: 2304.11771 [econ.GN]. URL: https://arxiv.org/abs/2304.11771.

[24] Yupeng Chang et al. *A Survey on Evaluation of Large Language Models*. 2023. arXiv: 2307.03109 [cs.CL]. URL: https://arxiv.org/abs/2307.03109.

[25] J Clusmann et al. "The future landscape of large language models in medicine." In: *Communications Medicine* 3.1 (2023), pp. 141–141.

[26] Christopher J Halbrook et al. "Pancreatic cancer: Advances and challenges". In: *Cell* 186.8 (2023), pp. 1729–1754.

[27] Jason Hoelscher-Obermaier et al. *Detecting Edit Failures In Large Language Models: An Improved Specificity Benchmark*. 2023. arXiv: 2305.17553 [cs.CL]. URL: https://arxiv.org/abs/2305.17553.

[28] Ayesha Juhi et al. "The capability of ChatGPT in predicting and explaining common drug-drug interactions". In: *Cureus* 15.3 (2023).

[29] Fenglin Liu et al. "A medical multimodal large language model for future pandemics". In: *NPJ Digital Medicine* 6.1 (2023), p. 226.

[30] Marissa Mock et al. "AI can help to speed up drug discovery—but only if we give it the right data". In: *Nature* 621.7979 (2023), pp. 467–470.

[31] Yujia Qin et al. *ToolLLM: Facilitating Large Language Models to Master 16000+ Real-world APIs*. 2023. arXiv: 2307.16789 [cs.AI]. URL: https://arxiv.org/abs/2307.16789.

[32] Vipula Rawte, Amit Sheth, and Amitava Das. *A Survey of Hallucination in Large Foundation Models*. 2023. arXiv: 2309.05922 [cs.AI]. URL: https://arxiv.org/abs/2309.05922.

[33] Yongliang Shen et al. *HuggingGPT: Solving AI Tasks with ChatGPT and its Friends in Hugging Face*. 2023. arXiv: 2303.17580 [cs.CL]. URL: https://arxiv.org/abs/2303.17580.

[34] Karan Singhal et al. "Large language models encode clinical knowledge". In: *Nature* 620.7972 (2023), pp. 172–180.

[35] Xiangru Tang et al. "Medagents: Large language models as collaborators for zero-shot medical reasoning". In: *arXiv preprint arXiv:2311.10537* (2023).

[36] Shitao Xiao et al. *C-Pack: Packaged Resources To Advance General Chinese Embedding*. 2023. arXiv: 2309.07597 [cs.CL].

[37] Yunzhi Yao et al. *Editing Large Language Models: Problems, Methods, and Opportunities*. 2023. arXiv: 2305.13172 [cs.CL]. URL: https://arxiv.org/abs/2305.13172.

[38] Geyan Ye et al. *DrugAssist: A Large Language Model for Molecule Optimization*. 2023. arXiv: 2401.10334 [q-bio.QM]. URL: https://arxiv.org/abs/2401.10334.

[39] Wayne Xin Zhao et al. *A Survey of Large Language Models*. 2023. arXiv: 2303.18223 [cs.CL]. URL: https://arxiv.org/abs/2303.18223.

[40] Angels Balaguer et al. *RAG vs Fine-tuning: Pipelines, Tradeoffs, and a Case Study on Agriculture*. 2024. arXiv: 2401.08406 [cs.CL]. URL: https://arxiv.org/abs/2401.08406.

[41] Elijah Berberette, Jack Hutchins, and Amir Sadovnik. *Redefining "Hallucination" in LLMs: Towards a psychology-informed framework for mitigating misinformation*. 2024. arXiv: 2402.01769 [cs.CL]. URL: https://arxiv.org/abs/2402.01769.

[42] Kevin Zheyuan Cui et al. "The Productivity Effects of Generative AI: Evidence from a Field Experiment with GitHub Copilot". In: *SSRN* (2024).

[43] Yunfan Gao et al. *Retrieval-Augmented Generation for Large Language Models: A Survey*. 2024. arXiv: 2312.10997 [cs.CL]. URL: https://arxiv.org/abs/2312.10997.

[44] Zorik Gekhman et al. *Does Fine-Tuning LLMs on New Knowledge Encourage Hallucinations?* 2024. arXiv: 2405.05904 [cs.CL]. URL: https://arxiv.org/abs/2405.05904.

[45] Qiao Jin et al. "GeneGPT: augmenting large language models with domain tools for improved access to biomedical information". In: *Bioinformatics* 40.2 (Feb. 2024). Ed. by Jonathan Wren. ISSN: 1367-4811. DOI: 10.1093/bioinformatics/btae075. URL: http://dx.doi.org/10.1093/bioinformatics/btae075.

[46] Haoqiang Kang, Juntong Ni, and Huaxiu Yao. *Ever: Mitigating Hallucination in Large Language Models through Real-Time Verification and Rectification*. 2024. arXiv: 2311.09114 [cs.CL]. URL: https://arxiv.org/abs/2311.09114.

[47] Jon M. Laurent et al. *LAB-Bench: Measuring Capabilities of Language Models for Biology Research*. 2024. arXiv: 2407.10362 [cs.AI]. URL: https://arxiv.org/abs/2407.10362.

[48] Shervin Minaee et al. *Large Language Models: A Survey*. 2024. arXiv: 2402.06196 [cs.CL]. URL: https://arxiv.org/abs/2402.06196.

[49] Oded Ovadia et al. *Fine-Tuning or Retrieval? Comparing Knowledge Injection in LLMs*. 2024. arXiv: 2312.05934.

[50] Mayk Caldas Ramos, Christopher J. Collison, and Andrew D. White. *A Review of Large Language Models and Autonomous Agents in Chemistry*. 2024. arXiv: 2407.01603 [cs.LG]. URL: https://arxiv.org/abs/2407.01603.

[51] S. M Towhidul Islam Tonmoy et al. *A Comprehensive Survey of Hallucination Mitigation Techniques in Large Language Models*. 2024. arXiv: 2401.01313 [cs.CL]. URL: https://arxiv.org/abs/2401.01313.

[52] Jinge Wang et al. "Bioinformatics and biomedical informatics with ChatGPT: Year one review". In: *Quantitative Biology* (2024).

[53] Lei Wang et al. "A survey on large language model based autonomous agents". In: *Frontiers of Computer Science* 18.6 (Mar. 2024). ISSN: 2095-2236. DOI: 10.1007/s11704-024-40231-1. URL: http://dx.doi.org/10.1007/s11704-024-40231-1.

[54] Hongjian Zhou et al. *A Survey of Large Language Models in Medicine: Progress, Application, and Challenge*. 2024. arXiv: 2311.05112 [cs.CL]. URL: https://arxiv.org/abs/2311.05112.

[55] ChromaDB. *ChromaDB*. URL: https://github.com/chroma-core/chroma.